# Best Practices for Convolutional Neural Networks Applied to Object Recognition in Images

**Anderson de Andrade**
Department of Computer Science
University of Toronto
Toronto, ON M5S 2E4
adeandrade@cs.toronto.edu

## Abstract

This research project studies the impact of convolutional neural networks (CNN) in image classification tasks. We explore different architectures and training configurations with the use of ReLUs, Nesterov's accelerated gradient, dropout and maxout networks. We work with the CIFAR-10 [15] dataset as part of a Kaggle competition [8] to identify objects in images. Initial results show that CNNs outperform our baseline by acting as invariant feature detectors. Comparisons between different preprocessing procedures show better results for global contrast normalization and ZCA whitening. ReLUs are much faster than tanh units and outperform sigmoids. We provide extensive details about our training hyperparameters, providing intuition for their selection that could help enhance learning in similar situations. We design 4 models of convolutional neural networks that explore characteristics such as depth, number of feature maps, size and overlap of kernels, pooling regions, and different subsampling techniques. Results favor models of moderate depth that use an extensive number of parameters in both convolutional and dense layers. Maxout networks are able to outperform rectifiers on some models but introduce too much noise as the complexity of the fully-connected layers increases. The final discussion explains our results and provides additional techniques that could improve performance.

## 1 Introduction

One of the main concerns in image classification is dealing with input invariances. Changes in size, position, background and angle of the objects inside images are difficult to account for. As a result, many algorithms rely on cleverly hand-engineered features to represent the underlying data. This preprocessing step requires expensive computations and limit the potential accuracy that a training algorithm can achieve by reducing the dimensionality of the feature space. Using low-level data representations such as raw pixels with minimal preprocessing and no feature detectors would require a sufficiently large training set to learn the appropriate invariances by example.

Convolutinal networks are able to automatically extract features that are resilient to invariances by the use of three mechanisms: local receptive fields, weight sharing and subsampling. Local receptive fields are connected to overlapping patches of the image. When forcing these connections to share the same weights, they become a feature map.

Having several feature maps allow us to look for different patterns at different locations of the input image. During training, these patterns become pertinent to the classification task. Sharing weights over feature maps make them insensitive to small invariances. If the input image is shifted, the activations of the feature map will be shifted by the same amount but would otherwise be unchanged.

Subsampling feature maps reduces complexity of upper layers and provides additional translation invariance. The receptive fields of this type of layers are usually chosen to be contiguous and non-overlapping. In this way, the response of a unit in the subsampling layer will be relatively insensitive to small shifts of the image in the corresponding regions of the input space.

The architecture design of the convolutional layers has a great impact in performance. It can be composed of several pairs of convolutional and sub-sampling layers that are finally attached to a traditional fully-connected, fully-adaptive neural network. Every pair of layers in a convolutional network offers a larger degree of invariance to input transformations compared to the previous layer. The number of layers in the fully-connected neural network can increase to offer more model complexity.

Convolutional networks are usually trained using the backpropagation algorithm. Training neural networks in general is an open research problem [1]. The use of non-linear activation functions introduces complexity in the objective function, making it difficult to optimize. Also, stacking hidden layers of units with these non-linear activation functions introduces difficulties during backpropagation. As stated in [1], at the beginning of learning, units in the upper layers will saturate and prevent gradients to flow backward, stopping learning in the lower layers.

The usually high number of parameters in neural networks allows models to easily overfit to the training set and do not generalize well to testing samples. Introducing regularization methods to solve these consequences without penalizing accuracy is a major concern. Different techniques have been suggested in [1]. Here, we explore the dropout technique, in which a fraction of the units in the neural network are randomly deactivated in each feedforward and backpropagation iteration. This prevents the units from co-adapting too much. Dropout training is similar to bagging, where many different models are trained on different subsets of the data. In this method however, each model is trained for only one step and all the models share the same parameters.

In [6], it is suggested that the best performance from dropout may be obtained by directly designing an architecture that enhances its abilities as a model averaging technique. Maxout networks learn not just the relationship between hidden units, but also the activation function of each hidden unit. This research project compares both regularization techniques.

## 2 Research objective

Our research focus on improving accuracy in image classification. The dataset is available as part of a Kaggle competition [8] and measures performance with the percentage of labels that are predicted correctly. Currently the best result is of 91.82%. We want to analyze and test different techniques to improve the baseline and attain similar state-of-the-art results.

We decided to use the cross-entropy error function instead of the sum-of-squares for this classification problem, since it leads to faster training as well as improved generalization [2]. Formally, for multiclass classification, we will use softmax outputs with the multi-class cross-entropy function given by:

$$E(\mathbf{w}) = -\sum_{n=1}^{N}\sum_{k=1}^{K} t_{kn} \ln y_k(x_n, \mathbf{w})$$

Where $n$ indexes the training samples $N$, $k$ indexes the classes $K$, the target variables $t_k \in \{0, 1\}$ follow the 1-of-$K$ coding scheme indicating the class, and the network outputs $y_k$ is a function of the input $x_n$ and the weight vector $\mathbf{w}$.

## 3 Dataset

To perform all our experiments we use the CIFAR-10 dataset [15], which comprises 60,000 32x32 color images in 10 classes, with 6,000 images per class. We will use 50,000 samples split 90-10 into training and validation sets respectively. The 10 classes belong to the following objects: airplane, automobile, bird, cat, deer, dog, frog, horse, ship, and truck.

To understand the complexity required in our classification models, we plotted the most representative two dimensions after performing PCA. A comparison between the CIFAR-10 and the MNIST dataset can be seen in Figure 1. Although the two most representative features are used, there is no clear boundary between any class in CIFAR-10, whereas with MNIST, it is easier to discern the boundaries between them. It becomes apparent that classifying objects is a much more difficult task than handwritten digits.

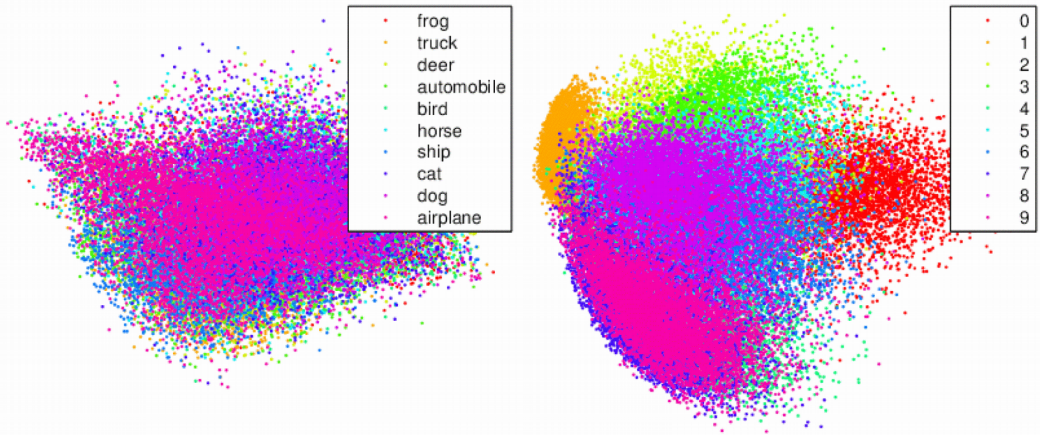

Figure 1: scatter plot of the two most relevant dimensions of the CIFAR-10 and MNIST datasets on the left and right respectively. Boundaries for CIFAR-10 are more difficult to obtain.

## 4  Baseline and initial results

A single-layer neural network with no convolutional layers was designed as our baseline. This model is useful to analyze the impact these layers have in feature extraction. The single hidden layer has 1,000 units with sigmoid activation functions. Following [5], RGB raw pixel values were used as feature matrices after rescaling them to [0,1] and centering its dimensions around zero.

This model was trained using stochastic gradient descent with mini-batches of size 100, a learning rate of 0.12, and classical momentum [3] with a viscosity (velocity) of 0.9. Figure 2 shows learning curves for this model trained for 30 epochs with no regularization.

Initial implementations of a convolutional neural network show promising results. The design employed 2 stacks of convolutional and sub-sampling layer pairs. The first convolutional layer has 6 feature maps and a kernel size of 5x5. The second convolutional layer doubles the feature maps to 12 while using the same kernel size. In both sub-sampling layers, max-pooling is done over a 2x2 region with no overlap. The model uses a single hidden fully-connected layer and all units have sigmoid activation functions, except for the outputs. Training was done with stochastic gradient descent with mini-batches of size 100, no momentum, and a learning rate of 1. For preprocessing, the same procedure as with the baseline model was performed, but in addition the images were converted to grayscale.

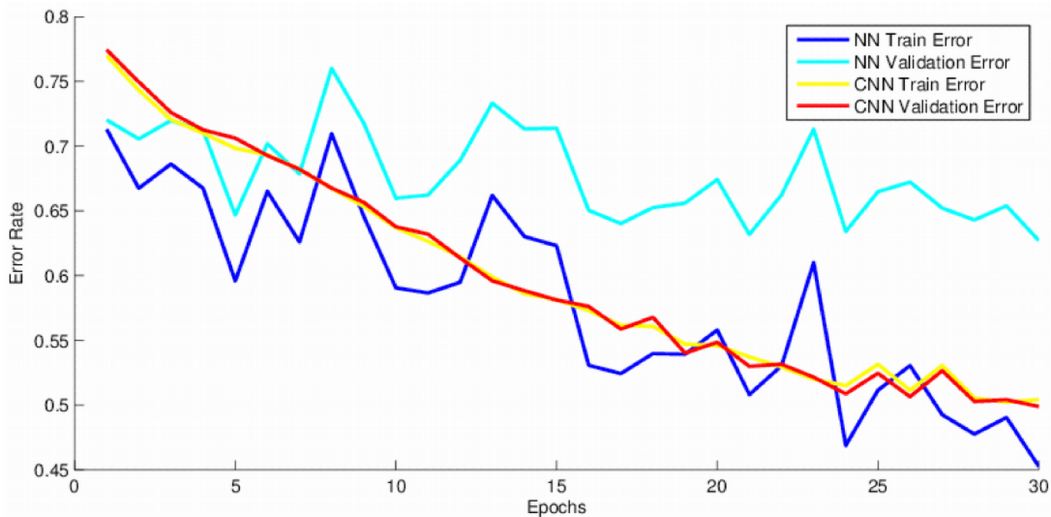

Figure 2: learning curves for the 1-hidden-layer neural network and convolutional network. The latter reaches better results and doesn't seem to converge after 30 epochs.

Figure 2 compares the results of our baseline and convolutional network over 30 epochs with no regularization or early stopping. The convolutional network and baseline attain 49.90% and 64.02% accuracy on the validation set respectively. Although the training curves reach similar values, the validation curve of the convolutional network model is able to follow its training curve very closely and get a 14.12% improvement over the baseline. These results justify our research.

## 5 Methodology and final results

### 5.1 Preprocessing

For preprocessing, the aim is to keep the procedure very simple and features as low-level as possible. We will limit ourselves to rescaling, centering, global contrast normalization (GCN), and ZCA whitening [9] of the features. To speed computations, a simple approach is to convert the images to a single grayscale channel. We could also simplify dimensions by learning feature representations using spherical K-means following [4]. The procedure consists of extracting pixel patches from the images in the training set and running K-means. Then, to project any input datum into the new feature space, it uses the thresholded matrix-product of the datum with the centroid locations, given by the activation function $z$:

$$z = max\{0, |D^T x| - \alpha\}$$

Where $D$ is the normalized dictionary, $x$ is the input and $\alpha$ is a hyperparameter to be chosen. Many choices of unsupervised learning algorithms are available for feature learning, such as autoencoders, RBMs, and sparse coding [1]. However, this variant of K-means clustering is widely used and according to [7], can yield results comparable to these other methods while also being simpler and faster.

Performing learning in the whole image can offer a significant performance boost. Since the images in the dataset are relatively small (32x32 pixels), we decided to avoid the explained method above to get better results. In this section only, we recur to grayscale conversions to speed computations.

The preprocessing step in [5], which attained state-of-the-art results on the LSVRC-2010 dataset, only involved rescaling and centering raw pixels. However, a common practice in image classification [6] is to perform global contrast normalization and ZCA whitening.

#### 5.1.1 Results

Figure 3 shows a performance comparison of different approaches using the initial architecture described in Section 4. For ZCA whitening we used a fudge factor of 0.01. The best values on the validation set are attained by using GCN and ZCA whitening with no rescaling. Ignoring the method using the same procedure with rescaling, this configuration yields a 2.88% improvement over the next one and is also the fastest. After several iterations the training errors go up presumably because of a slightly high learning rate. Further experiments will only use this preprocessing technique.

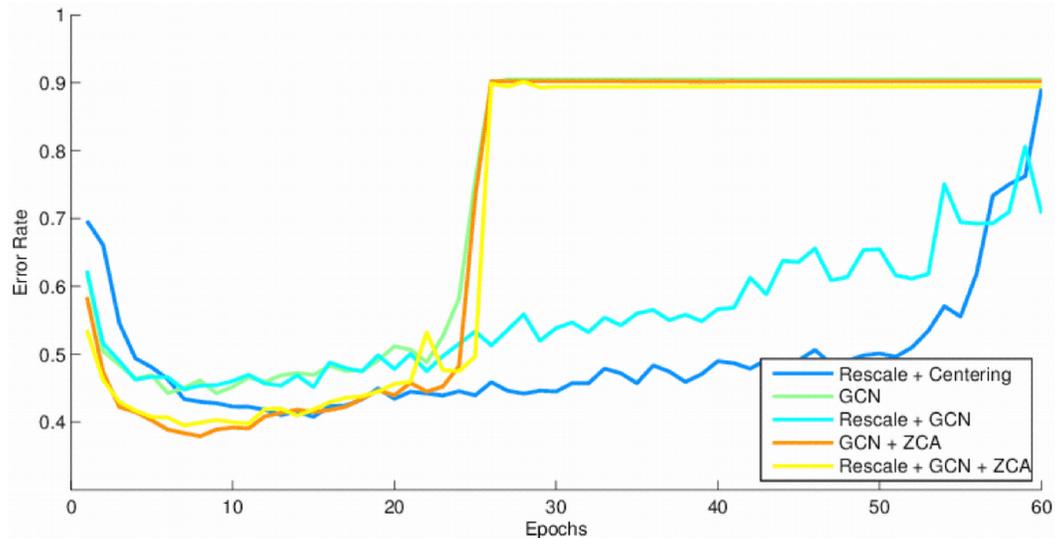

Figure 3: learning curves of the validation set for a combination of different preprocessing techniques. GCN with ZCA whitening attains the best results faster.

### 5.2 Implementation

Although we relied on our own implementation to develop our baseline results and preprocessing benchmarks, it wasn't really capable of taking advantage of parallel computations through GPUs. If our intention is to achieve state-of-the-art results we need to design bigger neural networks with many more parameters than the previous models. For dropout to work well, it also requires highly parametrized models that can easily overfit the data.

Training a model under such conditions would take an immense amount of time on CPUs. Therefor, we made extensive use of Pylearn2 [13], a machine learning framework developed in Python. It builds on top of Theano [14], a numerical computation library that compiles to run efficiently on GPU architectures such as CUDA.

We had at our disposition 8 Nvidia Tesla Fermi M2090 cGPUs, with 6GB of RAM each, providing enough memory to fit our largest neural network. The application server uses version 4.3 of the CUDA libraries. With this set-up, training our largest model took 2 days.

### 5.3 Activation units

Units using sigmoid or tanh functions have been known for saturating easily during training, preventing gradients to flow back to other layers. Even though convolutional layers have a reduced number of units due to parameter sharing, our aim is to introduce a considerable number of these layers combined with multiple dense, fully-connected layers. With this configuration, it is thought sigmoid functions will perform poorly. The recently suggested Rectified Linear Units (ReLUs), a non-saturating function given by $z = max(0, x)$, seem to train several times faster in [5].

#### 5.3.1 Results

Figure 4 shows a comparison between the three types of units using a deep model very similar to [5], involving a convolutional-pooling pair layer followed by two successive convolutional layers, a max-pooling layer plus two dense hidden layers. ReLUs are faster, and they seem to reach better results for the small number of epochs used. The use of ReLUs in all following experiments is justified.

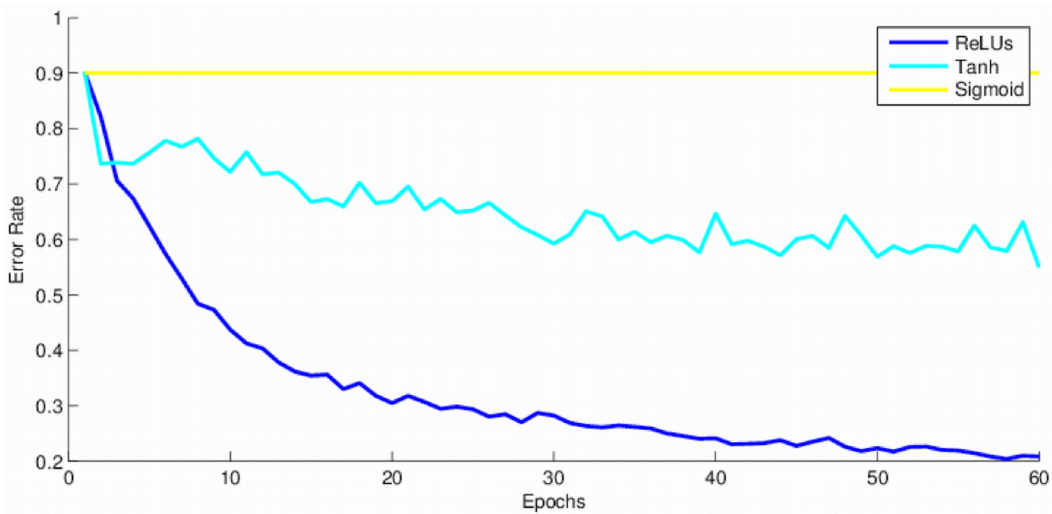

Figure 4: learning curves of the validation set for CNN models using units with ReLU, tanh and sigmoid activation functions.

## 5.4 Training

In the aims of providing a fair comparison between the models and regularization techniques being presented, the same training schedule and hyperparameters were used for all experiments. Optimization is addressed by using Nesterov's accelerated gradient with mini-batches [3]. The viscosity (velocity) and weight updates are given by:

$v_{t+1} = \mu v_t - \epsilon \Delta f(\theta_t + \mu v_t)$

$\theta_{t+1} = \theta_t + v_{t+1}$

Where $\epsilon > 0$ is the learning rate, $\mu \in [0, 1]$ is the momentum coefficient, and $\Delta f(\theta_t)$ is the gradient of the parameters at time $t$. It is common [5, 10] to use a fixed momentum of 0.9, or to initialize it at 0.5 and increase it gradually during the first 10-20 epochs until 0.9.

We found out that after increasing momentum to values close to 0.7, the negative log-likelihood in our maxout networks starts going up until it reaches intolerable values. The noise provided by dropout and maxout networks allows the training algorithm to explore different regions of the objective function with a large learning rate [6, 12]. Momentum introduces a weighted direction based on the history of previous steps. Setting a large weight for that direction in this chaotic environment seems to be affecting maxout units negatively. Even in dropout, although the model is still able to learn, better results were attained with a lower momentum. It seems that the same learning rate gives maxout networks a more oscillatory behavior than dropout networks. Hence, best results were attained when momentum was initialized to 0.5 and linearly increased until 0.6 after 250 epochs. This configuration will be used with the two regularization techniques.

As per our discussion, a sufficiently large learning rate of 0.17 was chosen to allow exploration of the objective function. To enforce the learning algorithm to eventually settle for a path and exploit it, we implement a linear decay of the learning rate, saturating in 500 epochs with a factor of 0.01 the original value. Since feature maps in convolutional networks share weights across different regions and see more training examples in an iteration, their weights should take smaller steps towards the gradient to prevent overfitting. Therefore, during training, gradients of the weights and biases in convolutional layers are scaled down to 0.05 times the global learning rate.

Although dropout and maxout act as powerful regularizers, they can't prevent the weights from reaching computational intractable values. We apply constrains to the weights by using the max-norm regularization [12], in which $||w_i|| \leq c$. In convolutional layers $w_i$ represents the weights of kernel $i$, whereas in a full-connected layer, $w_i$ is a vector of the weights incident on hidden unit $i$. The hyperparameter $c$ was set to $\sqrt{15/4}$, initially in all instances. However, a few models were still having overflow issues because of the weights in the first layer. Setting $c = 0.9$ for the set of kernel weights in that layer fixed the problem.

Weight initialization is even more crucial for dropout than maxout networks. Initial configurations got learning stuck in local minima for almost every model, requiring us to increase the range of the interval. After better settings were found, faster learning in the first iterations was attained by setting different sampling ranges according to the type of layer. Convolutional layers are more sensible to initialization and require a wider sampling range. We initialize their pertaining weights to $U(-0.5, 0.5)$, where function $U$ denotes the range of a uniform continuous distribution. Once there, initial training is faster if fully-connected layers are initialized with $U(-0.05, 0.05)$. Biases are initialized to 0.

Finally, we trained with mini-batches of size 100 and early stopping was used to abort training if the validation error does not improve for a good amount of epochs. The previous 20 misclassification errors on the validation set are kept. The algorithm will stop when the earliest value is the lowest in the set, returning the best version of the neural network found. There was no cap on the number of epochs.

## 5.5 Architectures

We experiment with different network architectures and study their impact on the classification task. There are many possible configurations but not many work well. As stated in [10], adding feature detectors increases the number of parameters and although adding pooling layers reduces the number of nodes in the next layer and introduces invariance, successive subsampling operations lose information about the position of the features in the image. As it was also stated in [10], pooling the maximum value (max-pooling) instead of the average usually works better in practice.

Taking in consideration these suggestions and observing other successful architectures, we decided to implement 4 models similar to those in LeNet-5 [11], deep convolutional neural networks [5], dropout in convolutional networks as in [12], and maxout networks [6]. The models are described as follows:

- **Model 1:** this model follows LeNet-5, but scales up the number of feature maps in similar proportions. It starts with 64 5x5 feature maps max-pooled to 2x2 with no overlap, followed by a convolutional layer of 96 5x5 feature maps using the same subsampling configuration. The following convolutional layer extracts 160 5x5 feature maps with no pooling. Unlike The original LeNet-5 [11], our single dense layer has 1,000 units. This model is characterized by an increasing number of feature maps of the same kernel size in each successive convolutional layer.

- **Model 2:** similar to [10] but reduces the kernel and pool sizes to accommodate for the lack of padding, this model has 3 convolutional layers with max-pooling. The first and second layer have 96 and 192 feature maps respectively with 5x5 kernels and a pooling region of 3x3 with a stride of 2, allowing for overlap. The third convolutional layer has 192 feature maps with 3x3 kernels and a 2x2 pooling region with the same stride 2 (no overlap). The model has only one fully-connected layer with 500 units. It explores having layers with large number of feature maps but with big pooling windows that effectively reduce the number of parameters in the upper layers.

- **Model 3:** the convolutional layers in this model are less parametrized than Model 2 and have a similar distribution of feature maps as Model 1. We incorporate overlapping pooling and add two large fully-connected hidden layers. The model is very similar to [12] but the size of the pooling window are reduced in the first layer to account for the lack of padding. The three convolutional layers have 5x5 kernels with 64, 64 and 128 feature maps respectively. The first layer has a 2x2 pool shape with a stride of 1. Same as Model 2, layers 2 and 3 have a pooling region of 3x3 with a stride of 2. The following two dense layers have 3,072 and 2,048 units, in order, from input to output. This model explores the impact of having a deep and large fully-connected neural network in the upper layers.

- **Model 4:** follows a similar architecture to that in [5] used for CIFAR-100. Since the dataset used here has less classes and the input images are smaller, the number of parameters is downscaled for each layer. The model has 5 convolutional layers and 2 fully-connected. The connections between convolutional layers 3-4 and 4-5 are not subsampled. Connections between 1-2, 2-3 and 5-6 are max-pooled with a region of 2x2 and a stride of 1, allowing again for some overlapping. As we stack convolutional layers we decrease the size of the kernels. Going from convolutional layers to fully connected layers, the number of feature maps are 32, 48, 64, 64, 48, and the kernel shapes are 8x8, 5x5, 3x3, 3x3, 3x3. The dense layers have 500 units each. This model explores depth in the number of layers. We reduced the number of parameters to make it comparable to other models. Even though it has 5 convolutional layers, supsampling is only performed 3 times as it goes up the network.

As empirically observed in [5], models with overlapping pooling are more difficult to overfit. All kernels have a stride of 1 so patches overlap as much as possible. Also, all models except Model 1 perform subsampling three times across the network. Only Model 1 does perform subsampling twice and so it retains more spatial information about the detected features in the image. All models use ReLUs and biases in the convolutional layers are shared across RGB channels.

### 5.5.1 Results

Figure 5 compares the performance of the 4 models, all trained under the same conditions presented in Section 5.4. The best learning curve belongs to Model 3 without maxout, achieving a 16.55% error rate on the validation set, though the best value is attained by Model 2 with maxout, with 16.26%. Model 4 obtains the worst performance with 22.97%.

Because of the performance obtained with Model 4, it can be said that depth is not such an important factor in obtaining good accuracy. Also, the fewer subsampling layers in Model 1 intended to retain spatial information, are not essential for performance.

Model 2 and Model 3 seem to represent the most reasonable architectures, suggesting that the number of parameters dominates performance. Model 2 has a lot of parameters (feature maps) in the convolutional layers while Model 3 has fewer feature maps, but adds an extra fully-connected layer with a higher number of units. Although the pooling windows in the first layer of Model 3 are smaller (2x2 with a stride of 1), the difference is not as considerable as the distribution of parameters between the convolutional and the fully-connected layers. Model 3 attained similar performance with considerably less feature maps just by adding a large fully-connected layer.

## 5.6 Regularization

In dropout [12], the choice of which units to drop is random. In the simplest case, all units are retained with a fixed probability. This probability is sampled independently for each hidden unit and for each training case. Since every configuration of activated units represents a different model, the predictions have to be averaged at test time. We could sample predictions using dropout and average the results by the number of models seen but this may require many samples to minimize error. An average approximation that does not require sampling and works well in practice multiplies the outgoing weights of that unit with its probability of being retained.

In our experiments, we could have explored different probability settings for different layers but we decided to use a common approach [12] where the probability of retaining units in the first layers is higher. Thus, the models have 0.8 probability for the input layer and 0.5 for the layers in the rest of the network.

From our discussion in 1.4, maxout networks introduces an activation function $h$ given by:

$$h_i(\mathbf{x}) = max_{j \in [1,k]} a_{ij}$$

Where $a_{ij}$ is the inner product of the input of the layer $\mathbf{x}$ and the weight vector $W_{ij}$ plus some bias $b_{ij}$:

$$a_{ij} = \mathbf{x}^T W_{ij} + b_{ij}$$

Unlike conventional neural networks where the activation function is applied to each $a_{ij}$, maxout networks group together $k$ units of $a_{ij}$ and take the maximum. In a convolutional layer, a maxout feature map can be constructed by grouping $k$ affine feature maps, usually, and as it is the case in our experiments, by channel and region. Following [6], a single maxout unit can be interpreted as making a piecewise linear approximation to an arbitrary convex function.

Different grouping schemes are employed depending on the type of layer. Convolutional layers use 2 linear pieces ($a_{ij}$) for a maxout unit and hidden layers use 5. These groups have strides equal to the number of pieces, which makes them disjoint. Maxout networks are considered a natural companion to dropout [6] and our experiments use the same dropout configuration explained above.

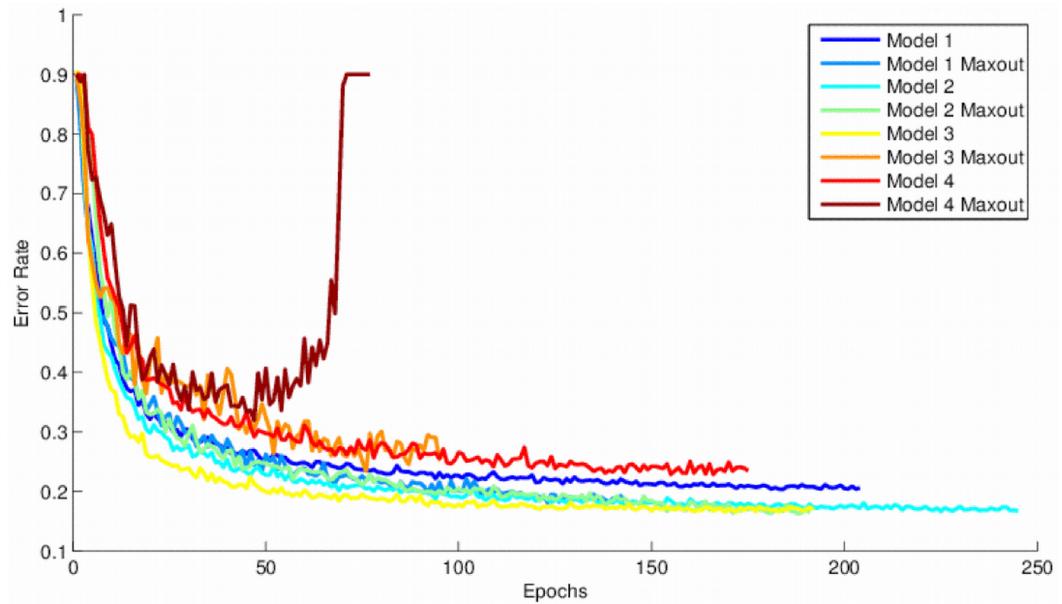

Figure 5: learning curves of the validation set for the four CNN models, trained under dropout and maxout with dropout.

### 5.6.1 Results

Shown in Figure 5, maxout improves the accuracy of Models 1 and 2 by 2.21% and 0.40% respectively, whereas Model 3 is considerably worse. Model 4 is halted by the early stopping algorithm in multiple runs but the curve is always above the dropout counterpart.

Although not shown in Figure 5, the gap between the training and validation curves with maxout seem to be smaller. Also, the learning curves in Models 3 and 4 have many oscillations. Both observations suggest that maxout is a stronger regularizer and introduces more noise than just using dropout. We hypothesize that the additional fully-connected layer in models 3 and 4 generates a more complex objective surface that the maxout method is not able to properly optimize because it is introducing too much noise. A solution could be to reduce the learning rate or make clusters assigned to maxout units in the fully-connected layers smaller. The current selected number of pieces for these layers was 5, it may have been set too high.

A hypothesis on why maxout may be a better architecture is that it better propagates information to the lower layers [6]. Even when a maxout unit is 0, the gradient information still flows to the parameters of the function, its units. Another reason entertained in [6] is that the max-pooling operation in convolutional layers fits more naturally as a part of maxout.

## 6 Conclusions

We laid down the foundations for our research and initial results showed that the endeavor was worth exploring. Indeed, convolutional neural networks are able to outperform fully-connected neural networks of similar complexity by learning invariant feature detectors. As we have also seen, ReLUs train much faster than tanh units and in the case of sigmoid units in can attain better results by avoiding saturation.

Although we cannot guarantee our preprocessing results can generalize to other datasets, we found that best results are obtained in CIFAR-10 when using GCN with ZCA whitening. We then designed and tested different models exploring different directions and found out that we attain better results by increasing the number of parameters over depth. The number of units in the fully-connected layers has as much impact as adding feature maps.

Maxout introduces more noise during training that can benefit accuracy. As complexity increases though, it can have a negative impact that can potentially be solved by using different training hyperparameters or more sparse maxout units. Our experiments imply that the method can't guarantee performance boosts in all situations.

There are many additional ideas we can explore to attain state-of-the-art results. First, padding the input image and the feature maps allows for better detection of features that may lie on the borders. Current state-of-the-art approaches seem to rely on this [5, 6].

Data augmentation techniques can reduce overfitting by artificially generating new training examples from the existent dataset [5]. These new examples posses some variance we hope the inference algorithm is able to account for. Some label-preserving transformations consist of generating image translations and horizontal reflections. A similar idea to padding would be to extract smaller patches of the image with random offsets. Altering the intensities of the RGB channels makes the learning algorithm invariant to small changes in color and illumination. This idea could be a more powerful alternative than preprocessing images with contrast normalization and whitening.

Lastly, it would be interesting to study the impact of different dropout and maxout configurations. We could, for instance, assign a higher probability of retaining units to convolutional layers and not just the inputs [5]. In maxout networks, the number of pieces per maxout unit can have a drastic impact in the complexity of the model that is worth studying.

**Acknowledgments**

We thank professor Brendan Frey for teaching many of the concepts involved in this project, the Department of Computer Science at University of Toronto for making their hardware and application servers available, and the Pylearn2 [13] and Theano [14] teams, whose blazingly fast algorithm implementations made computations feasible.